\documentclass[graybox]{svmult}

\usepackage{mathptmx}       % selects Times Roman as basic font
\usepackage{helvet}         % selects Helvetica as sans-serif font
\usepackage{courier}        % selects Courier as typewriter font
\usepackage{type1cm}        % activate if the above 3 fonts are
\usepackage{makeidx}         % allows index generation
\usepackage{graphicx}        % standard LaTeX graphics tool
\usepackage{multicol}        % used for the two-column index
\usepackage[bottom]{footmisc}% places footnotes at page bottom

\usepackage[numbers, square]{natbib}
\graphicspath{{./Figures/}}
\usepackage{color}
\usepackage{caption}
\usepackage{url, units}

\newcommand{\RC}[1]{{\sf #1}}
\newcommand{\todo}[1]{{#1}}

\usepackage[caption=false, font=footnotesize,subrefformat=parens,labelformat=parens]{subfig}
\usepackage{color}
\usepackage{units}

\makeindex             % used for the subject index
\begin{document}

\title*{From Formalised State Machines to Implementations of Robotic Controllers}
\author{Wei Li, Alvaro Miyazawa, Pedro Ribeiro, Ana Cavalcanti, Jim Woodcock \\ and Jon Timmis}
\institute{Wei Li and Jon Timmis \at Department of Electronics, University of York, Heslington, York, YO10 5DD, UK. \email{{wei.li, jon.timmis}@york.ac.uk}
\and Alvaro Miyazawa, Pedro Ribeiro, Ana Cavalcanti, and Jim Woodcock \at Department of Computer Science, University of York, Heslington, York, YO10 5GH, UK. \email{{ alvaro.miyazawa, pedro.ribeiro, ana.cavalcanti, jim.woodcock}@york.ac.uk}}
%From Formalised State Machines to Automatically Generated Code for Robot Control
\maketitle

%\abstract*{Each chapter should be preceded by an abstract (10--15 lines long) that summarizes the content. The abstract will appear \textit{online} at \url{www.SpringerLink.com} and be available with unrestricted access. This allows unregistered users to read the abstract as a teaser for the complete chapter. As a general rule the abstracts will not appear in the printed version of your book unless it is the style of your particular book or that of the series to which your book belongs.
%Please use the 'starred' version of the new Springer \texttt{abstract} command for typesetting the text of the online abstracts (cf. source file of this chapter template \texttt{abstract}) and include them with the source files of your manuscript. Use the plain \texttt{abstract} command if the abstract is also to appear in the printed version of the book.}

\abstract{Controllers for autonomous robotic systems can be specified using state machines. However, these are typically developed in an \textit{ad hoc} manner without formal semantics, which makes it difficult to analyse the controller. Simulations are often used during the development, but a rigorous connection between the designed controller and the implementation is often overlooked. This paper presents a state-machine based notation, RoboChart, together with a tool to automatically create code from the state machines, establishing a rigorous connection between specification  and implementation. In RoboChart, a robot's controller is specified either graphically or using a textual description language. The controller code for simulation is automatically generated through a direct mapping from the specification. We demonstrate our approach using two case studies (self-organized aggregation and swarm taxis) in swarm robotics. The simulations are presented using two different simulators showing the general applicability of our approach.}

%This paper presents a state machine-based framework for sound combination of notations for modelling and simulation, together with validation and verification techniques, to develop control systems for autonomous robots. 

\section{Introduction}\label{sec:introdution}

Safety is a major concern for autonomous robots, and the ability to provide evidence that a robotic system is safe can  be demanding. Formal verification is the process of checking whether a design satisfies some requirements (properties) or that an implementation conforms to a design, and it has been used to verify a variety of robotic systems such as service robots~\citep{Webster16} and swarming robots~\citep{Rouff07, Winfield05}.

Swarm robotics investigates how multiple robots, each with limited ability, communicate, coordinate and self-organize to accomplish certain tasks. Swarm robotics has potential in a wide range of real-world applications such as search and rescue, object transportation and environmental monitoring~\citep{Brambilla13}. While using a number of simple robots to collectively perform complex tasks is desirable, designing individual controllers to guarantee the emergence of certain swarm behaviour is challenging. If swarm robotic systems are to transfer from lab-based experiments to real applications, especially those that are safety-critical, the verification of the individual controllers as well as their resulting emergent swarm behaviours needs to be conducted in a rigorous way. 
%, as these two processes are usually conducted in isolation.
%in a \textit{ad hoc} manner without a rigorous connection between the specification defined in a formal way. This makes it hard to maintain the consistency between formal verification and the executive code deployed in the robots, as no rigorous connection between them is established. 

%TODO conclusion: proof the correctness of simulation code
Typically, the implementation of a robotic control system is conducted without establishing a strong connection between the controller code and the high-level design specifications. Here we explore the usage of a state-machine based notation, RoboChart~\citep{CWA16}, for designing robotic controllers. RoboChart has a formal semantics that allows for verification. In this paper, we extend RoboChart to support automatic code generation from the designed controllers to simulations. 
%the executable code for simulation is automatically generated through a direct mapping from the specification. 

%In this paper, the formal semantics and simulation code is a direct mapping from the model in RoboChart. The long-term objective is to mathematically proof the correctness of the mapping. 
%The tool will generate artefacts (e.g. analysis results and proofs) that can become a useful input for certification. 
%vIn the later case, the verification is narrowed down to only a specific robot, which makes it hard to be used in a wide variety of robots. 
%This means the connection between the high-level verification modelling language and low-level simulation or deployment is still lacking. 
%In the design stage, the state machine guides the development of a simulation or deployment, but no rigorous connection between them is established. The source code that is used for controlling robots are developed in a ad hoc manner. This makes the executable code hard to maintain and verify. 

%We explore it in use, with extra facilities to model time, probability and environment stimuli. The  is customised to make it suitable for verification and automatic code generation. 
%In the verification tools, the state machine controllers are specified using mathematical notions.

%TODO add one more reference of FSM to control single robot
%In addition, we also intend to model probability and environment stimuli. 
Finite state machines are often adopted to design robot controllers in swarm robotics~\citep{Liu10, Gauci14a, Gauci14b, Bjerknes13, Chen15}. A commonly used state-machine notation is that of UML~\citep{Bergenti00}. RoboChart takes inspiration from UML, and provides facilities to model timed and probabilistic systems, composed of one or more controllers. 

Formal verification has been investigated in the design of controllers in swarm robotic systems~\citep{ Rouff07, Winfield05, Dixon12, Konur12, Brambilla12}. In~\citep{Winfield05, Dixon12}, the authors used a temporal logic to formally specify and verify the emergent behaviour of a swarm robotic system performing aggregation. In~\citep{Konur12}, the authors used PRISM, a model checker for probabilistic automata, to formally verify the global behaviour of a foraging case scenario through exhausting all possible swarm behaviours. The analysis results were compared with those reported in~\citep{Liu10}, which used the test-driven simulation and showed a good correspondence. In these works, finite state-machine controllers were described using natural language, and there was no direct mapping from the high-level specification to low-level controller code. 

In~\citep{Lopes16}, the authors applied supervisory control theory to control a swarm of robots. Their approach supported automatic code generation. The controllers were specified using standard finite state machines, without any of the extra facilities for architectural modelling available, for example, in UML.
%however there was no support for modelling time and probability, and there was no formal semantics. 
%Through a collection of software tools, the properties of the control system can be automatically validated and verified. 

Various researchers have also explored the use of model-driven approaches to develop the high-level control of robots~\citep{SHLS09, DKSZZ12, Schultz07, Feiler12}. The
architecture analysis and design language (AADL) is a unifying component-based framework for modelling software systems with a particular focus on embedded real-time systems~\citep{Feiler12}. RoboChart could in principle be integrated into the controller component in AADL. In~\citep{Schultz07}, a language was developed to program self-assembling robots. They proposed a role-based language that allowed the programmer to define the behavioural roles of each component independently from the concrete physical structure of the robots. However, in these works, the controllers of robots (e.g. state machine) were not formally specified, which makes it difficult to reason about robotic systems.  
%generation of platform-independent code. It was defined as a UML profile. 

%In~\citep{Schultz07}, a domain-specific language was developed to program self-assembling robots. It is a rule-based language that allows the programmer to define the behavioural rules independently from the concrete physical structure of the robots.  The executive C code is automatically generated. 
%This model-driven approach for robotics has grown a large interest recently~\citep{SHLS09, DKSZZ12}. 
%RoboChart is also a domain-specific language used for designing robotic controllers. 

The main contribution of this paper is to reduce the gap between high-level specification and implementation of robotic controllers. 

This paper is organized as follows. Section~\ref{sec:robochart} briefly introduces RoboChart. This includes the elements of RoboChart and the approach to automatic code generation for simulation and deployment. Section~\ref{sec:modelling_using_robochart} presents two case studies (self-organized aggregation~\citep{Gauci14a} and swarm taxis~\citep{Bjerknes13}) in swarm robotics. The simulations using the automatically generated C++ code are presented. Section~\ref{sec:conclusion} concludes the paper and presents future work.

\section{RoboChart}\label{sec:robochart}
\begin{figure}[!t]
\centering
\includegraphics[width=0.7\textwidth]{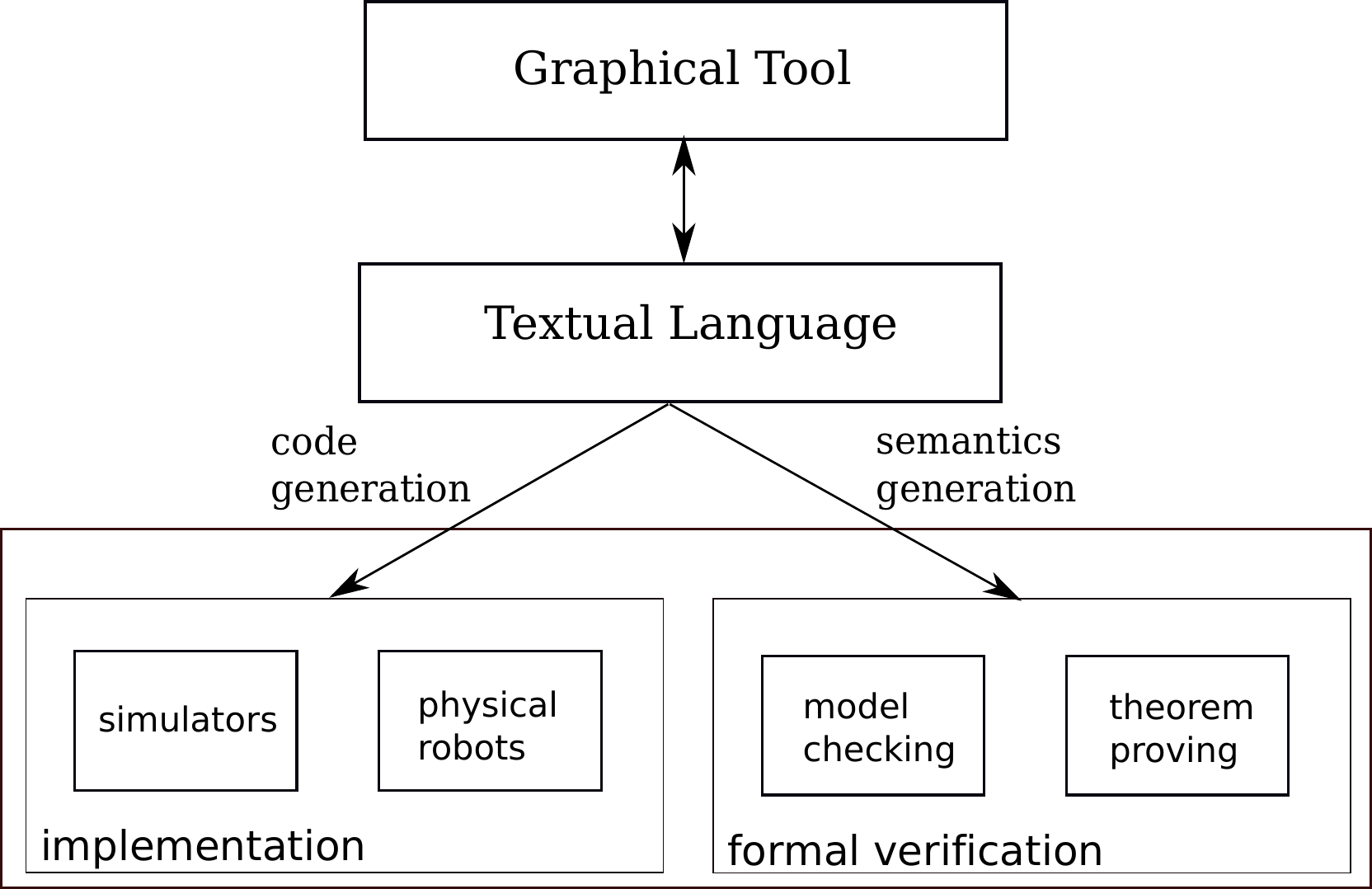}
\caption{The RoboChart framework for combining formalised state machines and implementation of robotic controllers.}
\label{fig:project_theme}
\end{figure}

%In this section, we present RoboChart, for the specification and design of robotic systems. Besides state machines, it also includes elements to organize specifications such as modules and controllers. 
%The state machine notation is fully specified, including specification for time properties. 
Figure~\ref{fig:project_theme} shows the RoboChart framework to combine formalised state machines and automatic implementation of robotic controllers. Once the controller is developed, code is generated automatically to be used in different simulation platforms or physical robots. Formal semantics are also automatically generated for verification. Details of the formal semantics of RoboChart can be found in~\citep{CWA16}. In the following section, we focus on the automatic code generation for simulation and deployment.

\subsection{Elements of RoboChart}

%Figure~\ref{metamodel:program} shows the structure of the RoboChart modelling tool. 
%RoboChart is composed of elements: modules, robotic platforms, controllers, state machines, interfaces, and types. Modules give a complete account of a robotic control system. They define robotic platforms or include references to platforms defined elsewhere to indicate the robots available. Modules associate their robotic platforms with particular controllers and state machines to specify the behaviours of the robots. A state machine can be directly used for controlling robotic platforms. 
%When the behaviour is complex, controller that encapsulates multiple (potential interacting) state machines can be used. 

Central to RoboChart is a state-machine notation. RoboChart machines include states and their {\it entry}, {\it during} and {\it exit} operations (actions), as well as transitions possibly triggered by events. The entry operation is executed when the robot enters a state, and followed by the execution of the during operation. When a transition is triggered, the exit operation of the source state is executed. If an action is associated with the transition, it is also executed before the state machine enters the target state. 

Operations and events of a state machine are described in an \textit{interface}. A state machine can \textit{requires} an interface. An operation can either be described without implementation or implemented by the user in a state-machine style. An operation can include a \textit{precondition} and a \textit{postcondition}. 
%When an operation is implemented in RoboChart, this operation \textit{provides} (part of) the interface. 

%\textcolor{red}{Apart from the user-defined operations, some commonly used operations such as \RC{MoveForward}, \RC{MoveClockwise} are part of RoboChart API and could be imported into a library in RoboChart.}

Variables can be defined in a state machine, an interface or an operation. Different data types (primitive or composite) can be defined. When the behaviour is complex, multiple (potentially interacting) state machines can be used. 

In addition to state machines, RoboChart also includes elements to organize specifications such as modules and robotic platforms~\citep{CWA16}. A module defines a system, including a robotic platform and associated controllers.  Each controller can be specified by one or more state machines.

RoboChart also includes time constraints. A \textit{clock} can be defined inside a state machine to record the instant in time \RC{\#T} in which a
transition is triggered. For example, the primitive \RC{since(T)} yields the time elapsed since the most recent time instant {\#T}. If \RC{since(T)} is used as a condition (guard) on a transition with no events, then the transition will be taken immediately once the guard is true. Unless time is specified, we assume an operation takes no (or a significantly small) time. 

For full details of RoboChart, refer to~\citep{CWA16}.  
% A similar primitive \RC{sinceEntry(s)} yields the time elapsed since entering a state.
%\textcolor{red}{In RoboChart, there are two types of state machine: one for controlling the robot (defined by \textit{stm}) and the other for representing an \textit{operation}.}
%In this paper, we will demonstrate the usage of \RC{since(T)}. 

% For example, the primitive \RC{wait(t)} is used to specify that the robot waits for \RC{t} time units in a particular state. 

%Another time primitive is \RC{e@t}, which records the time elapsed between the moment the event \RC{e} is available and when the transition it guards is triggered. The \RC{e@t} primitive could be used in the scenario that the robot needs to perform dead reckoning where the robot needs to know the time between entering the state and exiting the state. Other time primitives can be found in the online supplementary materials~\citep{}. 
%In this paper, we will demonstrate the usage of primitive \RC{since(T)}. 

%\begin{figure}[!t]
%\includegraphics[width=\textwidth]{program_metamodel.png}
%\caption{Structure of RoboChart modelling tool}
%\label{metamodel:program}
%\end{figure}

\subsection{Simulation and deployment}

In RoboChart, the robot's controller is specified either graphically or using a textual description language. The automatically generated controller code can be imported into a wide variety of simulation platforms. 
%Note that it is also possible to directly import the generated code onto a physical robot\textcolor{red}{yes but there will be constrains -maybe we need to be a little carefull here?}. 
%Examples are V-rep~\cite{KA04} and Webots~\cite{Mic04}, which are integrated development tools that can simulate different kinds of robots, simulation scenarios and sophisticated physics engines. 

%We use object-oriented simulations. This paradigm is adopted in many simulators and supports a direct mapping of RoboChart
%constructs~(controllers, state machines, and so on) to simulations in a well structured way.

%A module defines a complete simulation, although additional information about the environment and simulation configuration is needed.  

\begin{figure}[!t]\centering
  \includegraphics[width=0.55\textwidth]{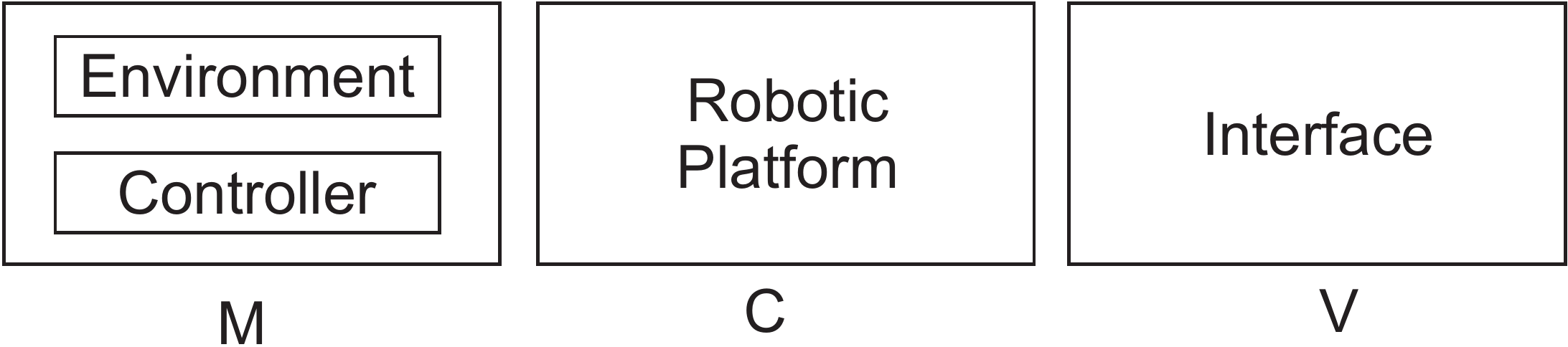}
  \vspace{0.1cm}
  \captionof{figure}{RoboChart simulations pattern}
  \label{figure:mvc}
\end{figure}

We adopt the model-view-controller~(MVC) pattern in the design of simulations, where, the terms model and controller are used in a different way from that adopted in RoboChart.
Figure~\ref{figure:mvc} maps the RoboChart constructs to an MVC architecture. The model~(M) component contains a simulation of the environment and of the RoboChart controller. We can generate a simulation of the RoboChart controller, potentially together with a simulation of the environment\footnote{The specification of environment is still under development. Currently the environmental stimuli are manually defined in the simulation.}.
The controller component~(C) implements the robotic platform, which corresponds to a particular robot in a simulation. Finally, the view component~(V) defines the visualisation of the simulation.

%It provides the variables, events, and operations defined in the RoboChart robotic platform. 
%We can generate a simulation of the RoboChart controller, which, together with a simulation of the environment defines the M component.

%With this approach, we have simulations that can take advantage of different simulations of robotic platforms and can be visualised in different ways, without requiring changes to the M component. For example, the environment model in the M component can be implemented with a variety of different physics engines, for example,
%Bullet\footnote{\url{http://bulletphysics.org/wordpress/}} or
%Delta3D\footnote{\url{http://delta3d.org/}}. Likewise, for the V component, we may choose to view the robot with different visualization tools in 3D, or 2D or in fact choose only a textual
%representation of the platform to allow for debugging. 

%\begin{table}[!t]
%\centering
%\begin{tabular}{|l|l|}
%	\hline
%	{\bf RoboChart} & {\bf Simulation}
%	\\ \hline %
%	states & attribute of enumerated type
%	\\ %
%	clock & attribute of timer class
%	\\ %
%	interface & inherit interface class
%	\\ %
%	\hline
%\end{tabular}
%	\vspace{2ex}
%	\caption{RoboChart state-machine classes}{\label{table:robochart-simulations} }
%\end{table}

\captionsetup[subfigure]{labelformat=empty}  
\begin{figure}[!t]
	\centering
	\subfloat
	{
\begin{tabular}{|l|l|}
	\hline
	{\bf RoboChart} & {\bf State machine class}
	\\ \hline %
	states & attribute of enumerated type
	\\ %
	clocks & attribute of timer class
	\\ %
	interfaces & inherit interface class
	\\ %
	\hline
\end{tabular}
	}
	\subfloat{
\begin{tabular}{|l|l|}
	\hline
	{\bf RoboChart} & {\bf Interface class}
	\\ \hline %
	events & attribute of enumerated type
	\\ %
	variables & attribute
	\\ %
	operations & methods
	\\ %
	\hline
\end{tabular}
	}
	\caption{RoboChart state machine and interface classes}
	\label{fig:state_machine_interface_class}
\end{figure}

We now describe how the controllers defined in RoboChart can be mapped into an executable language, specifically C++. Other object-oriented languages can be considered in a similar way, but are currently outside the scope of our work. The simulation of a controller is the simulation of its state machine(s).
Each machine is implemented by a class. If the machine requires an interface, that interface is also implemented by a class, which is inherited by the state machine. 

Figure~\ref{fig:state_machine_interface_class} defines how constructs of a state
machine and interface are mapped to elements of a class. The variables and events defined in an interface are generated as attributes of the class. The
operations ({\it entry}, {\it during} and {\it exit}) that the robot executes in a state give
rise to methods. We note that, even if an operation is specified in a state-machine style,
it is generated as a method. To update the state machine, some other methods such as \textit{MakeTransition} are also generated.

If a \textit{clock} is defined in a state machine, a timer class is generated. It has a attribute \textit{counter}, indicating the elapsed time, and methods such as \textit{StartTimer} and \textit{ResetTimer}. The state machine includes an object of the timer class as an attribute. The timer is used as a service of the state machine, which means the state machine can assess the counter. The state of the robot is updated in a cyclic manner, with the length of the cycle linked to the length of time required to capture events. The counter of the timer is updated in each control cycle.
% elapsed time and \textit{timerEnabled} indicating whether the timer is enabled or not
%The timer class has a variable counter and other variables to indicate whether the timer is enabled or it needs to trigger a wait condition.
%Note that if the interface only includes only constants, it is generated as a separate header file including all the variables rather than a class (shown in Section~\ref{sec:modelling_omega_algorithm}). 

A primitive data type is directly mapped into one in C++. For example, the type \textit{real} corresponds to \textit{double} in the code. A composite type is generated as a pre-defined class. For example, \textit{vector2d} corresponds to a 2D vector class. The data-type system in RoboChart as well as its mapping are still under development.
%The main state machine has the states and timer class (if any) as attributes and inherits the required interface classes. 

%The operations of the API have default implementations that assume that the
%preconditions hold.  This allows more efficient implementations. We plan to
%include the preconditions in the simulation code to support verification.

%In a simulation, the state of the robot is updated in a cycle-based way
%(that is, in control steps). Therefore, the time in a simulation is discrete
%and the length of the control step is the smallest granularity for a clock.
%%Note that the control step is simulation time, not the real time (e.g. time
%%measured by CPU clock).
%If the controller is implemented in a sequential way, we also need to assume
%that the length of the control step is small enough in order not to miss any
%events.

\section{Modelling robotic controllers using RoboChart}\label{sec:modelling_using_robochart}

To demonstrate our approach, we investigate two case studies on canonical problems in swarm robotics: aggregation~\citep{Gauci14a} and swarm taxis (flocking towards a beacon)~\citep{Bjerknes13}. In these case studies, the robots are homogeneous. The controller of each robot is defined by a single state machine, and it is executed in the e-puck~\citep{Francesco09}, which is a differential wheeled robot. It has an inter-wheel distance of $\unit[5.1]{cm}$. The maximum speed for the left and right wheels of the e-puck is $\unit[12.8]{\textrm{cm/s}}$, forward or backward.

\subsection{Case study one: aggregation}
\label{sec:case_studies}

\subsubsection{Aggregation behaviour}\label{sec:aggregation_behaviour}

In this behaviour, each robot is equipped with a line-of-sight sensor that \todo{detects the type of item in front of it}. The range of this sensor is unlimited in simulation. 
%We assume that there are $n$ types (e.g., background, other robot, object~\citep{Gauci14a, Gauci14b}). The state of the sensor is denoted by $I\in\{0,1,\ldots,n-1\}$.
%Given $n$ sensor states, \todo{the mapping} can be represented using $2n$ system parameters, which we denote as:
%\begin{equation}\label{controller:form}
%\mathbf{p} = (v_{\ell 0}, v_{r0}, v_{\ell1}, v_{r1}, \cdots, v_{\ell (n-1)}, v_{r (n-1)}).
%\end{equation}
%Using $\mathbf{p}$, any reactive behaviour for the above robot can be expressed. 
It gives a reading of $I=1$ if there is a robot in the line of sight, and $I=0$ otherwise. The environment is free of obstacles. The objective for the robots is to aggregate into a single compact cluster as fast as possible. 

Each robot implements a reactive behaviour by mapping the sensor input ($I$) onto the outputs, that is, a pair of predefined speeds for the left and right wheels,
$(v_{\ell I}, v_{rI})$, $v_{\ell I}, v_{rI} \in \left[-1,1\right]$,  where $-1$ and $1$ correspond to the wheel rotating backwards and forwards respectively with maximum speed. 

The parameters of the aggregation controller were found by performing a grid search over the space of possible combinations~\citep{Gauci14a}. The controller exhibiting the highest performance was:
\begin{equation}\label{eq:aggregation_optimal_controller}
\mathbf{p} = \left(v_{\ell0}, v_{r0}, v_{\ell1}, v_{r1}\right) = \left(-0.7, -1.0, 1.0, -1.0\right). 
\end{equation}

When $I=0$, a robot moves backwards along a clockwise circular trajectory with a linear speed of $\unit[-10.88]{cm/s}$ and an angular speed of $\unit[-0.75]{rad/s}$. When $I=1$, a robot rotates clockwise on the spot with a linear speed of $0$ and the maximum angular speed of $\unit[-5.02]{rad/s}$. 
%Note that rather counterintuitively, an robot never moves forward, regardless of $I$. 
%With this controller, a robot provably aggregates with another robot or a quasi-static cluster of robots~\citep{Gauci14a}.

\subsubsection{Modelling the aggregation controller in RoboChart}

\begin{figure}[!t]
	\centering
	\includegraphics[width=\textwidth]{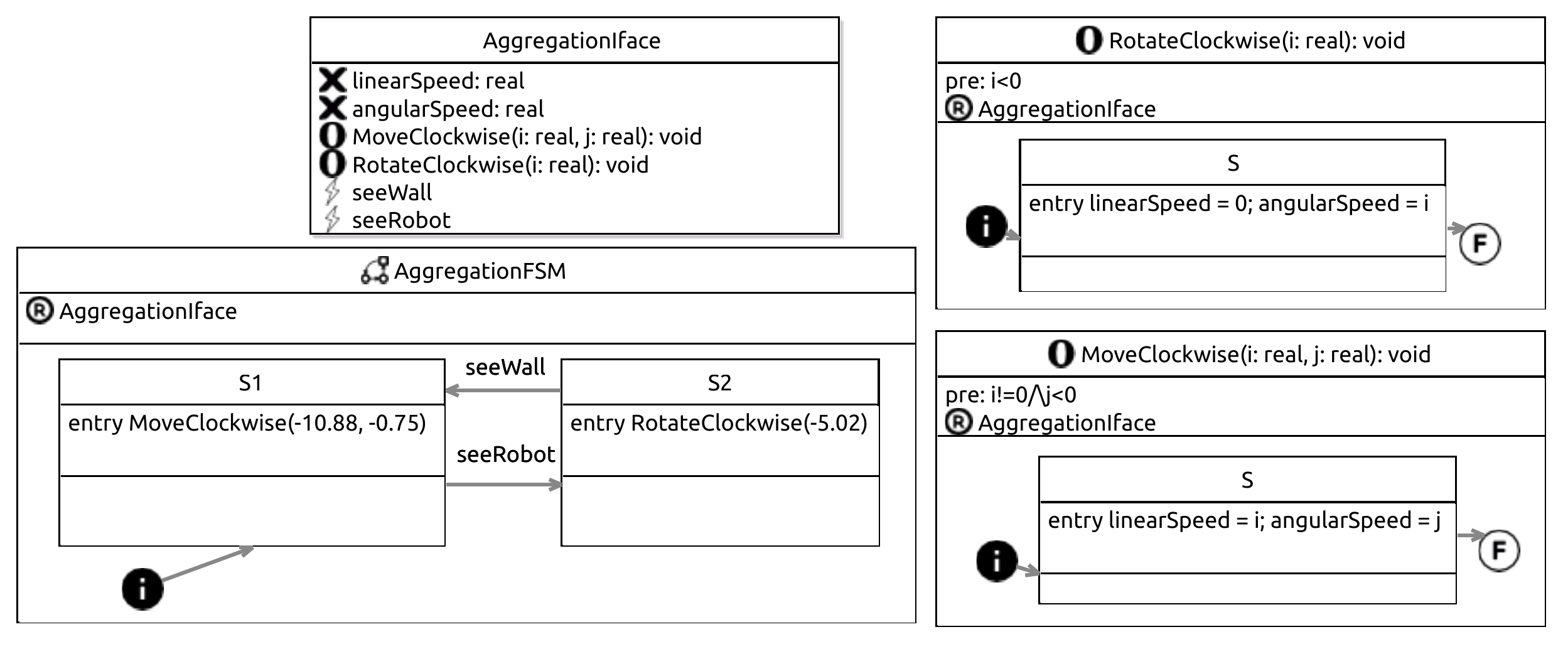}
	\caption{Diagram of the aggregation controller modelled in Robochart.}
	\label{fig:aggregation_controller}
\end{figure}

\begin{figure}[!t]
	\centering
	\subfloat[(a) aggregation controller]{  %\scriptsize{initial configuration}
		\includegraphics[height = 1.8 in]{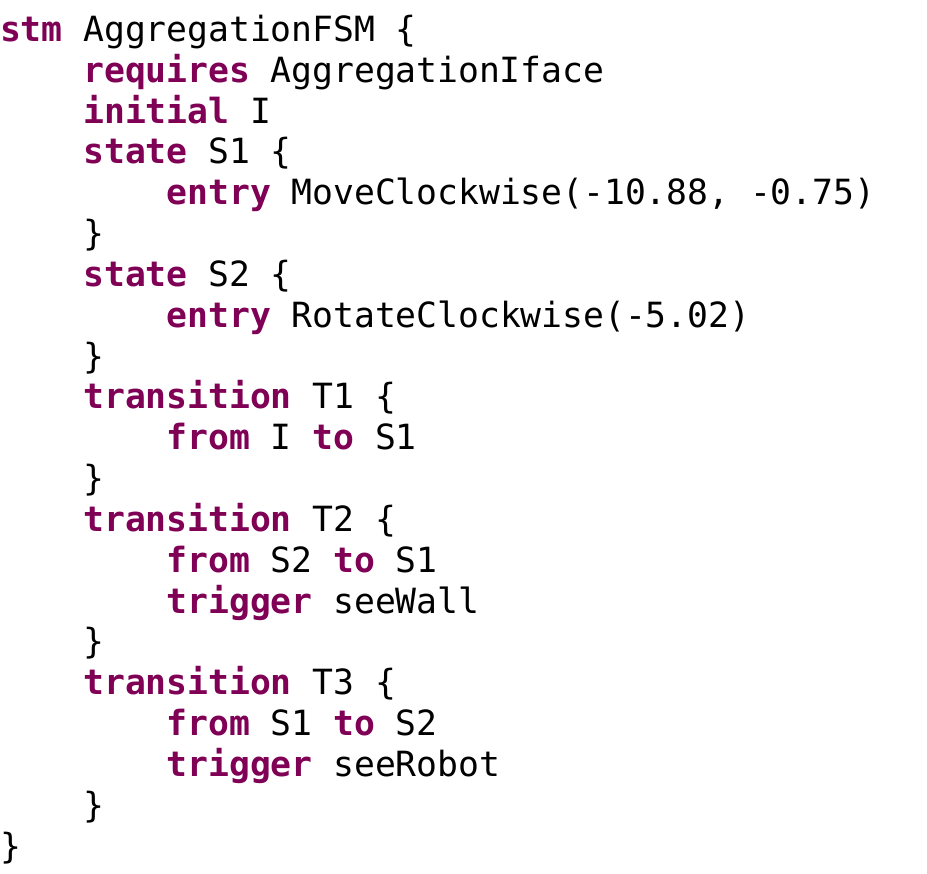}  %1.25
	}
	\subfloat[(b) \textit{MoveClockwise} operation]{
		\includegraphics[height = 1.8 in]{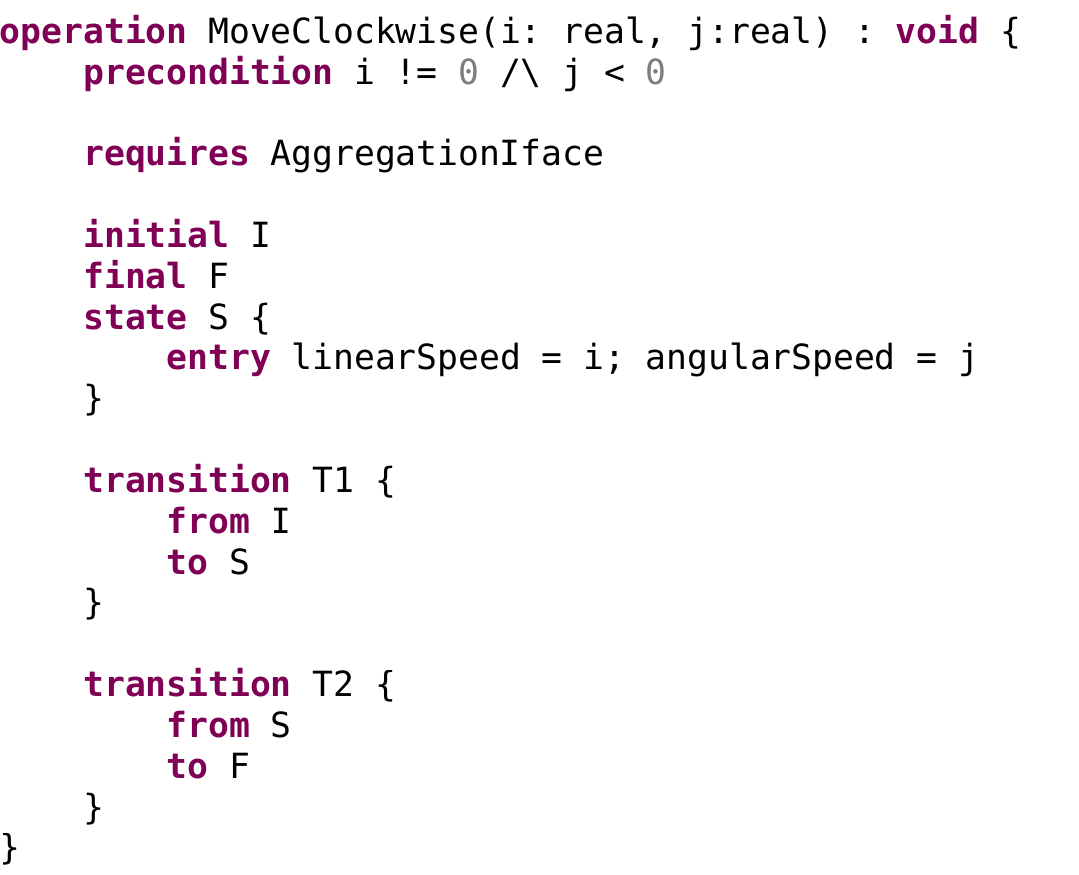}
	}
	\caption{Textual description of the aggregation controller and an operation in RoboChart.}
	\label{fig:aggregation_textual_description}
\end{figure}

Figure~\ref{fig:aggregation_controller} shows the diagram of the aggregation controller modelled in RoboChart. An interface, \RC{AggregationIface}, declares the variables, operations and events. The state machine (\RC{AggregationFSM}) requires \RC{AggregationIface}. The state machine has an initial node, \RC{i}, pointing to the initial state. The aggregation controller includes two states (\RC{S1} and \RC{S2}), two events (\textit{seeWall} and \textit{seeRobot}, which correspond to $I=0$ and $I=1$ respectively), and two operations (\textit{MoveClockwise} and \textit{RotateClockwise}). These operations are implemented in a state machine style with only an initial state \RC{S} and final state \RC{F}. Different from the \RC{AggregationFSM} state machine, both operations have a final state. An operation can include precondition that must be satisfied by the caller to guarantee that the functionality of this operation is realised as specified. For example, in the \textit{MoveClockwise} operation, the precondition requires that its first argument, an angular speed, is negative, and the second, the linear speed, is not zero. In the generated C++ code, this is realized using the \textit{assert} function. A textual description of the \RC{AggregationFSM} state machine and the \textit{MoveClockwise} operation is shown in Figure~\ref{fig:aggregation_textual_description}.
%To model a system, either the textual or graphical tool can be used, and the transformation between them is realized automatically. 
%The textual  can be extended to customize the simulation, for example, adding simulation configuration. 
%This means the operation terminates and finishes within a number of control steps\footnote{In the two case studies considered here, all the operations finish within one control step.}.
 
%TODO 
In the generated C++ code, two classes (\RC{AggregationIface} and \RC{AggregationFSM}) are generated. The class \RC{AggregationIface} includes the attributes of two double variables (\textit{linearSpeed} and \textit{angularSpeed}), two methods (\textit{MoveClockwise} and \textit{RotateClockwise}) and two boolean events (\textit{seeWall} and \textit{seeRobot}). The operations are generated as virtual functions that can be overridden if necessary. The \RC{AggregationIface} class is inherited by the state machine \RC{AggregationFSM} class. It has the attributes of states \RC{S1} and \RC{S2}, and other methods that are used to run the state machine. The generated C++ controller code can be found in the online supplementary materials~\citep{DARS2010Online}.

\subsubsection{Simulating the aggregation behaviour}

The automatically generated code of the aggregation controller is tested in Enki~\citep{Enki}, which has a built-in model of the e-puck robot. Enki is a 2D simulator and it can simulate swarms of robots a hundred times faster than real time. The speed of the left and right wheels of the e-puck can be set separately. The line-of-sight sensor in Enki is simulated by casting a ray from the e-puck's front and checking the first item with which it intersects (if any). The arena size is $\unit[250\times250]{cm^2}$, and the initial position and orientation of the robots are randomly distributed. The length of the control step is set to $\unit[0.1]{s}$, and the physics is updated every $\unit[0.01]{s}$.
%The automatically generated code of the robot's controller is tested in Enki~\citep{Enki} simulator. 
%The robot is represented as a disk of diameter $\unit[7.0]{cm}$ and mass $\unit[150]{g}$. The inter-wheel distance is $\unit[5.1]{cm}$. The speed of each wheel can be set independently. Enki induces noise on each wheel speed by multiplying the set value by a number in the range $(0.95, 1.05)$ chosen randomly with uniform distribution. The maximum speed of the e-puck is $\unit[12.8]{\textrm{cm/s}}$, forward or backward. 

\begin{figure}[!t]
	\centering
	\subfloat[initial configuration]{  %\scriptsize{initial configuration}
		\includegraphics[height = 0.7in, width = 1.1 in]{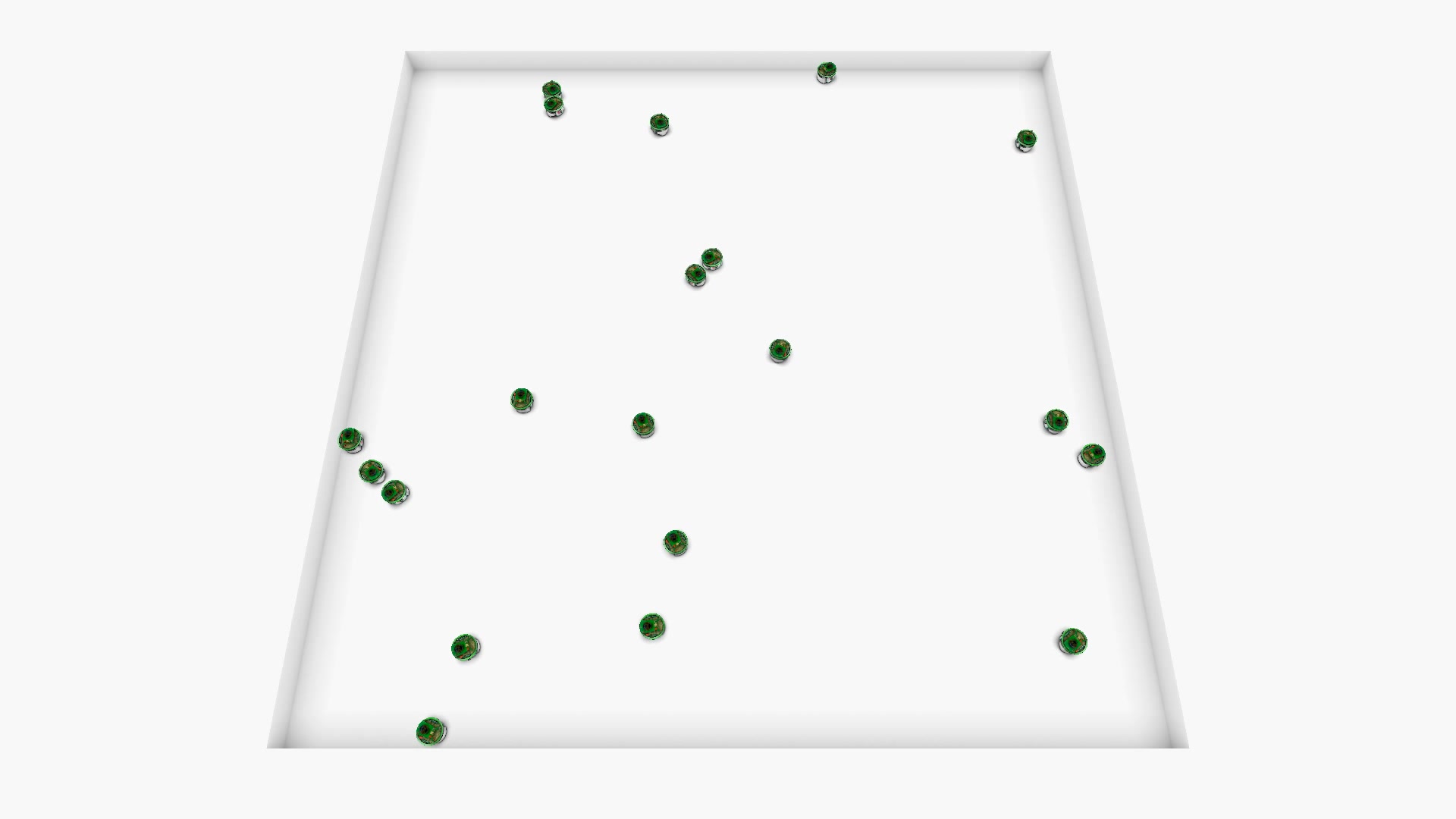}  %1.25
	}
	\subfloat[after $20$ $\unit{s}$]{
		\includegraphics[height = 0.7in, width = 1.1 in]{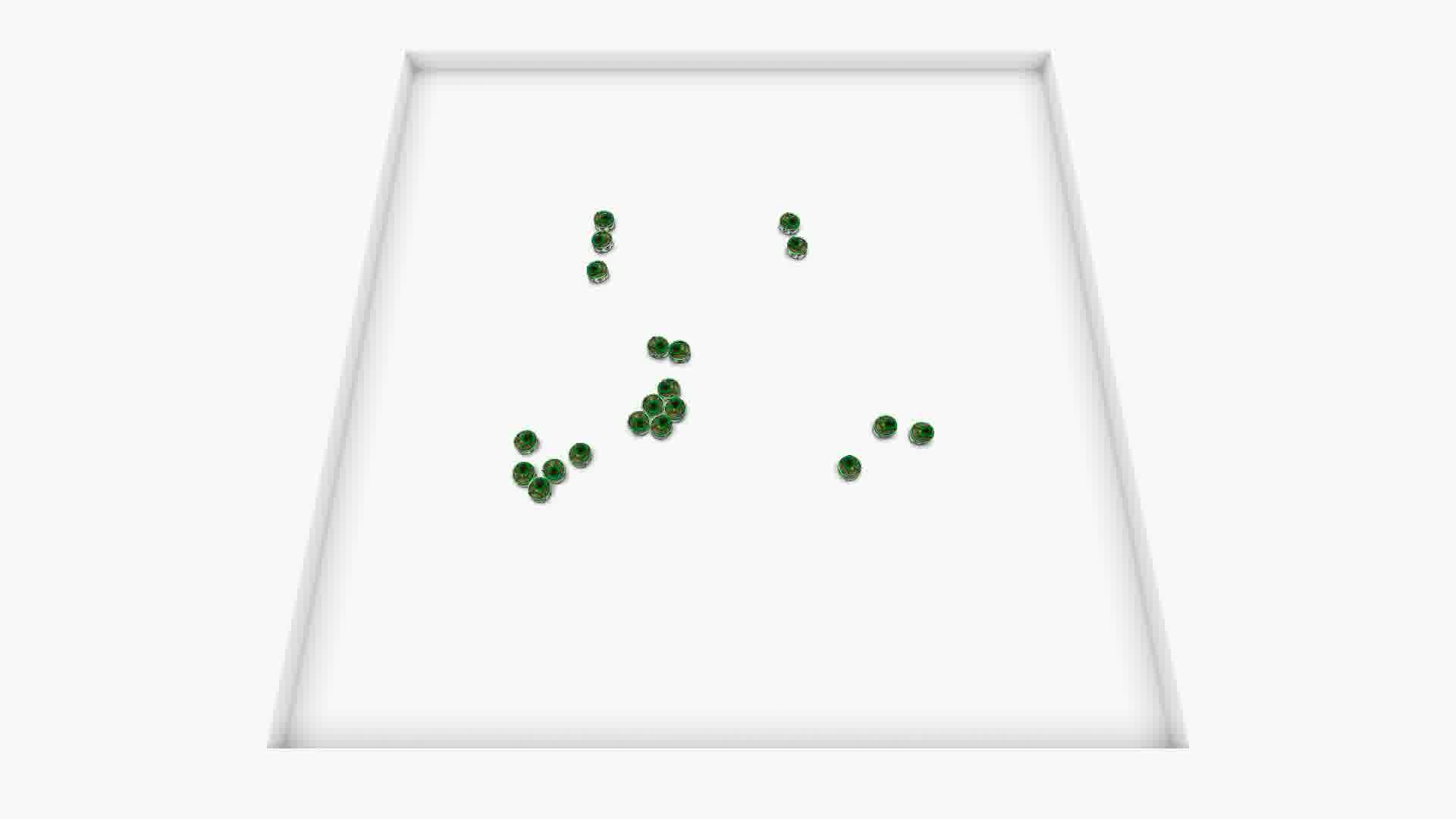}
	}
	\subfloat[after $40$ $\unit{s}$]{
		\includegraphics[height = 0.7in, width = 1.1 in]{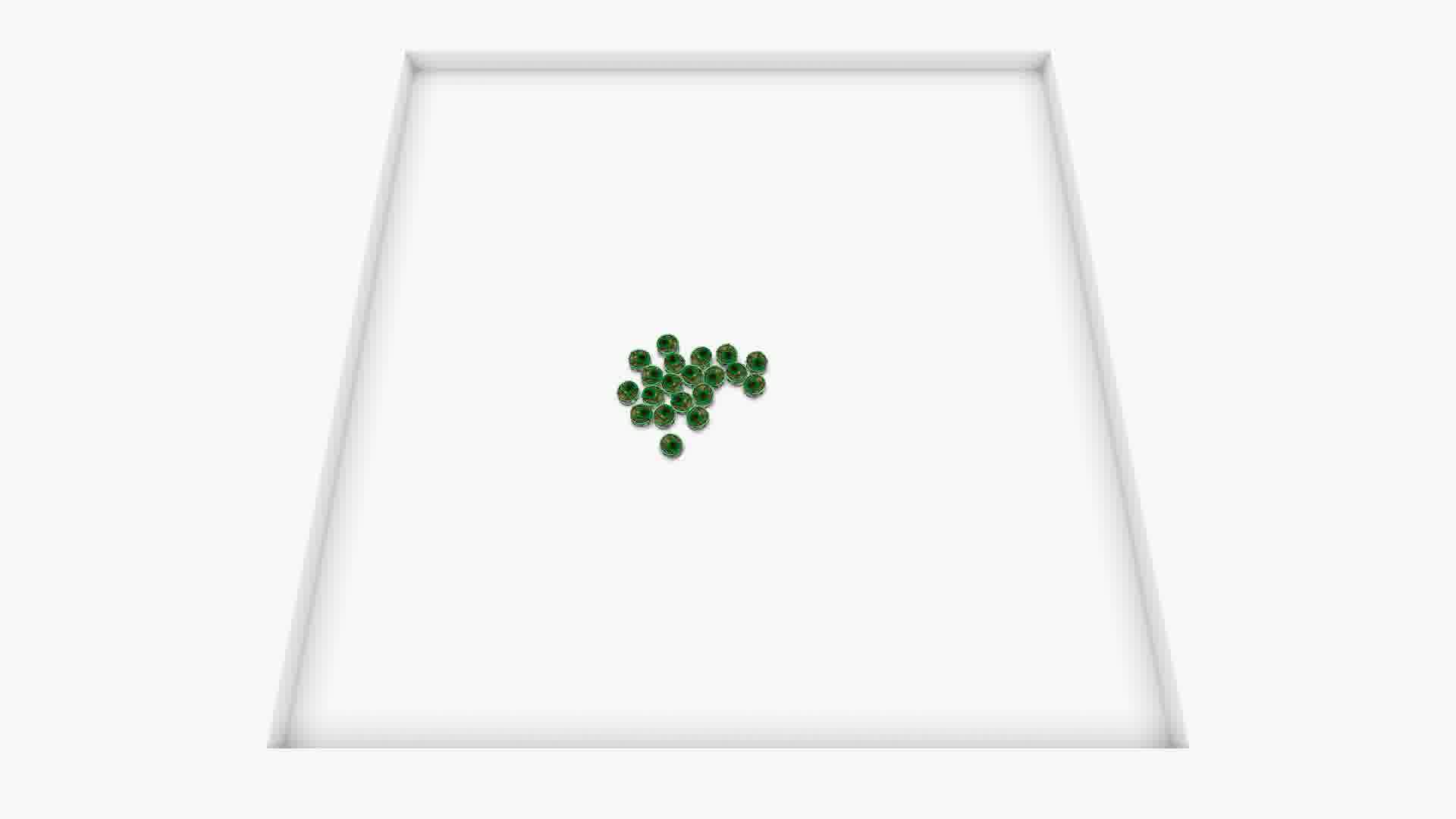}
	}
	\subfloat[after $60$ $\unit{s}$]{
		\includegraphics[height = 0.7in, width = 1.1 in]{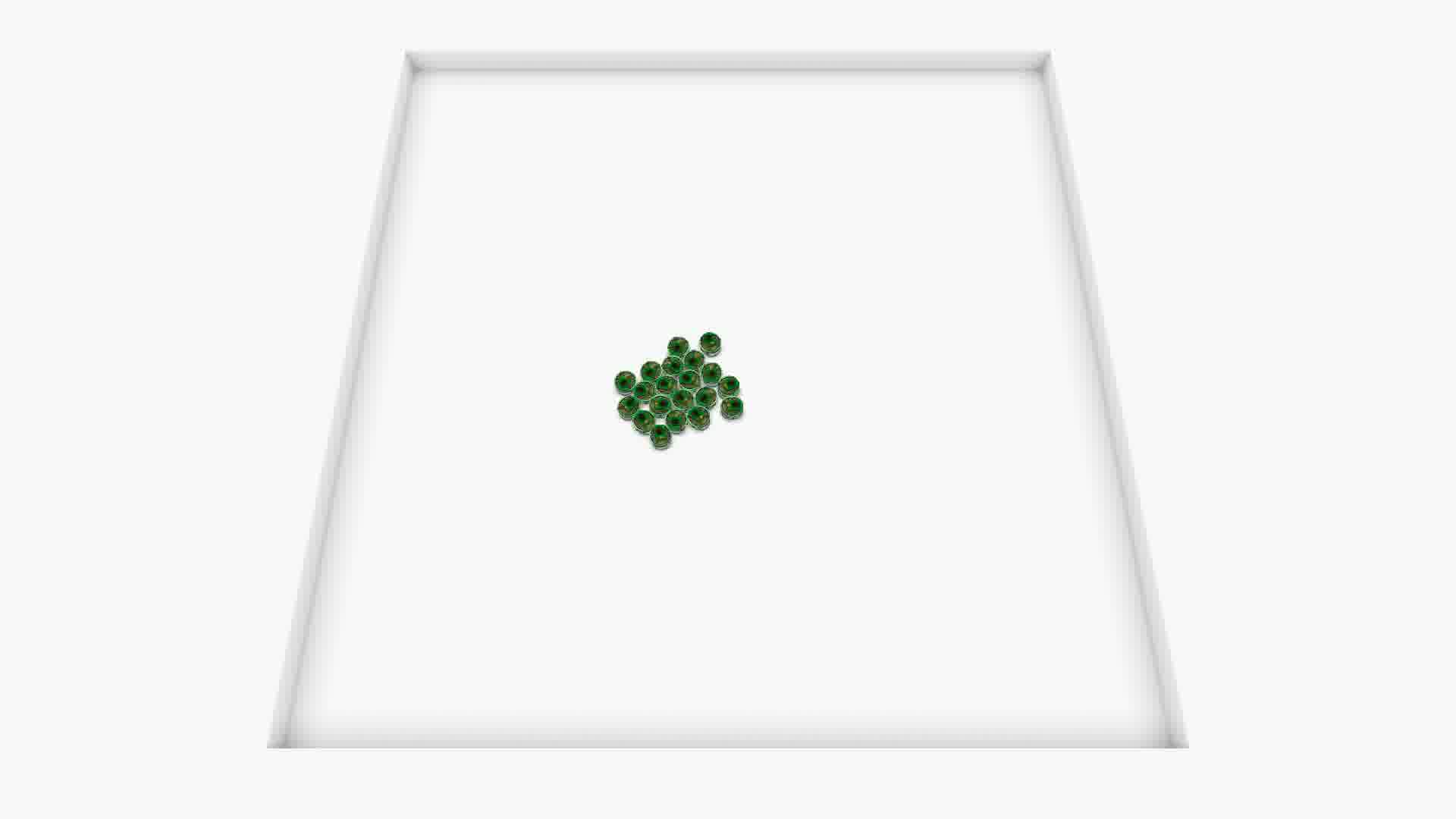}
	}
	\caption{Snapshots of the aggregation behaviour of $20$ robots in simulation, using the automatically generated controller code from the RoboChart model.}
	\label{fig:aggregation_snapshoot}
\end{figure}

We performed $10$ simulation trials with $20$ robots, and in each trial the robots can aggregate into a single cluster. Figure~\ref{fig:aggregation_snapshoot} shows snapshots from a simulation trial using the automatically generated controller code from the model in RoboChart. 
%The arena size is $200\times200$ square centimetres, and the robots are randomly distributed in the arena. 
%The implementation of operations that require the interface is generated in the scope of this interface.
%TODO Table~\ref{} shows the contents of \RC{Iface} and \RC{StateMachine} classes.
%The events are cleared after each run cycle of the main state machine. It also include a \RC{CheckInEvent} method that is overwritten by the user to update the events of the robot, for example, through checking some particular sensors.

\captionsetup[subfigure]{labelformat=empty}  

%\subsubsection{Object clustering}
%
%This behaviour uses the same line-of-sight sensor in Section~\ref{sec:aggregation_behaviour}, but has
%three sensor states: $I=0$ if the sensor is pointing at the background (e.g., the wall of the environment, if the latter is bounded), $I=1$ if the sensor is pointing at another robot, and $I=2$ if it is pointing at an object. The objective of the robots is to arrange the objects into a single compact cluster as fast as possible. 
%
%The controller's parameters, found using an evolutionary algorithm~\citep{Gauci14b}, are: $\mathbf{p} = \left( 0.5, 1.0, 0.1, 0.5, 1.0, 0.5 \right)$. 
%\begin{equation}\label{eq:clustering_optimal_controller}
%\mathbf{p} = \left( 0.5, 1.0, 0.1, 0.5, 1.0, 0.5 \right).
%\end{equation} 
%
%TODO replace the snapshots with the epuck robots in enki
\captionsetup[subfigure]{labelformat=empty}  

\subsection{Case study two: swarm taxis}
\label{sec:case_study_omega_algorithm}

%The omega algorithm consists of two \textcolor{red}{Need to be careful here, as taxis is not a behaviour in the defined sense, it is emergent.  Given we have been talking about behaviour based controllers, I think this is misleading}behaviours: flocking and swarm taxis towards a beacon~\citep{Bjerknes13}. 

\subsubsection{Swarm taxis behaviour}

In the swarm taxis behaviour, the robots move towards a beacon while maintaining a coherent group. Each robot has three states: \RC{Forward}, \RC{Coherence} and \RC{Avoidance}. The initial state is \RC{Forward}. If the robot is in the \RC{Forward} state for a certain number of time units without detecting any robots within avoidance radius, it enters the \RC{Coherence} state. In this state, the robot turns towards the estimated center of the nearby robots. If the robot detects any robot within the avoidance radius while it is in the \RC{Forward} state, it enters the \RC{Avoidance} state. In this state, the robot turns away from the estimated center of the robots being avoided. 
%The estimated center is calculated using the range and bearing sensors of the e-puck. 

The robot can be illuminated by a beacon in the environment or shadowed by other robots (unilluminated). The avoidance radius when the robot is illuminated is larger. The avoidance radius is updated while the robot is in the \RC{Forward} state. It is this mechanism that leads to the emergent swarm taxis  behaviour~\citep{Bjerknes13}.   
%The flocking behaviour is realized using \RC{Coherence} and \RC{Avoidance}.

\subsubsection{Modelling the swarm taxis controller in RoboChart}\label{sec:modelling_omega_algorithm}

\begin{figure}[!t]
	\centering
	\includegraphics[width=\textwidth]{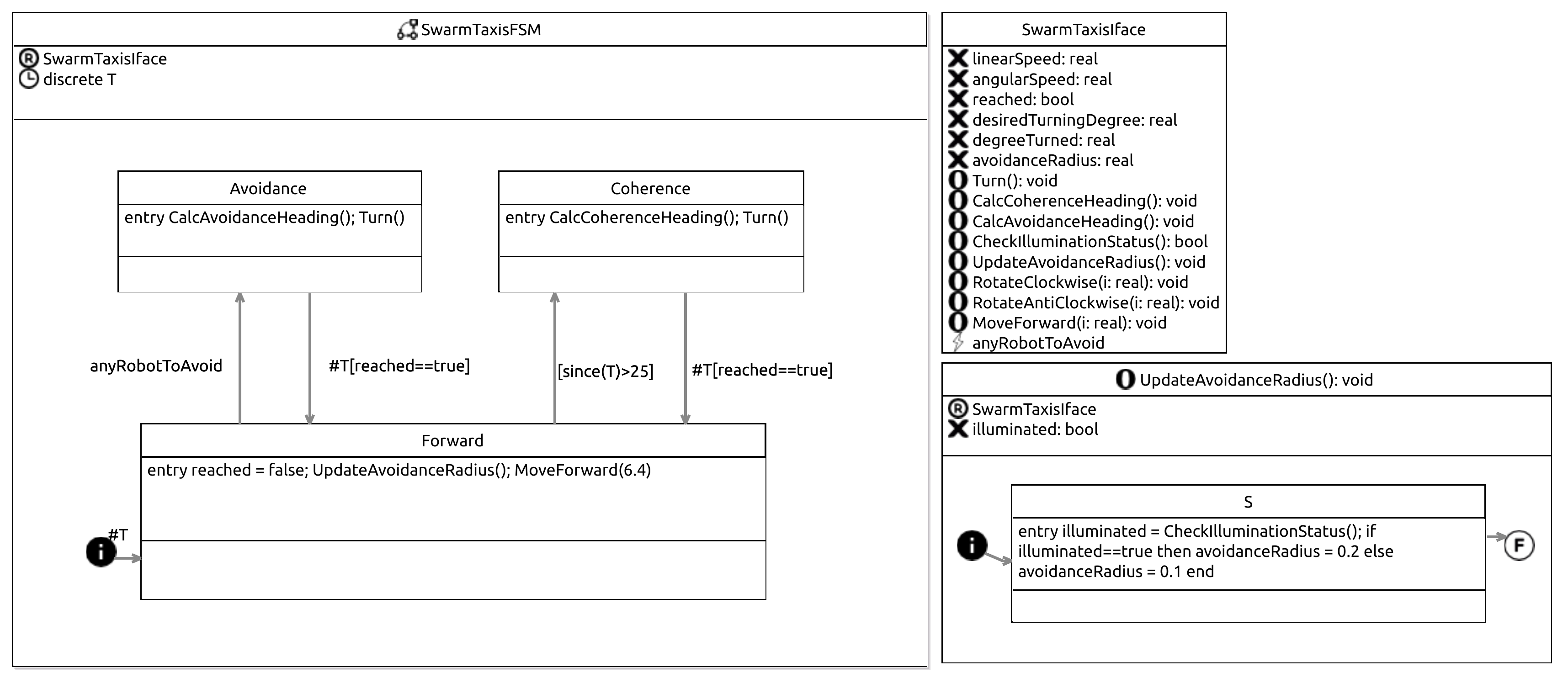}
	\caption{Model of the swarm taxis controller in Robochart.}
	\label{fig:omega_algorithm_controller_diagram}
\end{figure} 

Figure~\ref{fig:omega_algorithm_controller_diagram} shows the diagram of the swarm taxis controller in RoboChart. The full model can be found in the online supplementary materials~\citep{DARS2010Online}. The interface \RC{SwarmTaxisIface} defines the variables, operations and an event. A clock is defined inside the controller \RC{SwarmTaxisFSM}. The initial state of the controller is \RC{Forward}, where a timer \RC{T} is started immediately. The timer records the time the robot stays in the state \RC{Forward}. In RoboChart, an expression marked in square brackets (such as \textit{reached == true} or \textit{since(T) $<$ 25} in Figure~\ref{fig:omega_algorithm_controller_diagram}) is a guard for the transition. If there is no event associated with a transition, satisfaction of the condition will trigger the transition immediately. For example, once $25$ time units have elapsed since the robot is in the \RC{Forward} state, a transition from the \RC{Forward} state to the \RC{Coherence} state is triggered. 

In the \RC{Forward} state, the robot updates its avoidance radius through the operation \textit{UpdateAvoidanceRadius}. The actual avoidance radius is set based on the boolean variable \textit{illuminated} resulting from the operation \textit{CheckIlluminationStatus}. If the robot is illuminated, the avoidance radius is set to $0.2$; otherwise it is set to $0.1$. As a consequence of this choice, the robots that have longer avoidance radius (are illuminated) tend to move towards the beacon and thus give rise to the beacon taxis behaviour of the whole swarm. Note that although we have declared the operation \textit{CheckIlluminatedStatus}, we have chosen not to specify it in the RoboChart model, since it relies on the usage of the robot's sensors, which is platform dependent. If the robot detects any other robots nearby within the avoidance radius, it enters the \RC{Avoidance} state, where the robot calculates the desirable turning degree (\textit{desiredTurningDegree}) using the operation \textit{CalcAvoidanceHeading} and then executes the operation \textit{Turn}. In the operation \textit{Turn}, the boolean variable \textit{reached} is updated to indicate whether the robot has turned the desirable degree. Once the desirable turning degree has been achieved, the variable \textit{reached} is set to \textit{true}, which triggers the transition from \RC{Avoidance} to \RC{Forward}. Every time a transition is triggered, the timer \RC{T} is started. Similar operations occur in the transition from \RC{Coherence} to \RC{Forward}. 

For a full description of the model, refer to~\citep{DARS2010Online}.

\captionsetup[subfigure]{labelformat=empty}  
\begin{figure}[!t]
	\centering
	\subfloat[initial configuration]
	{
		\includegraphics[width = 1.1 in]{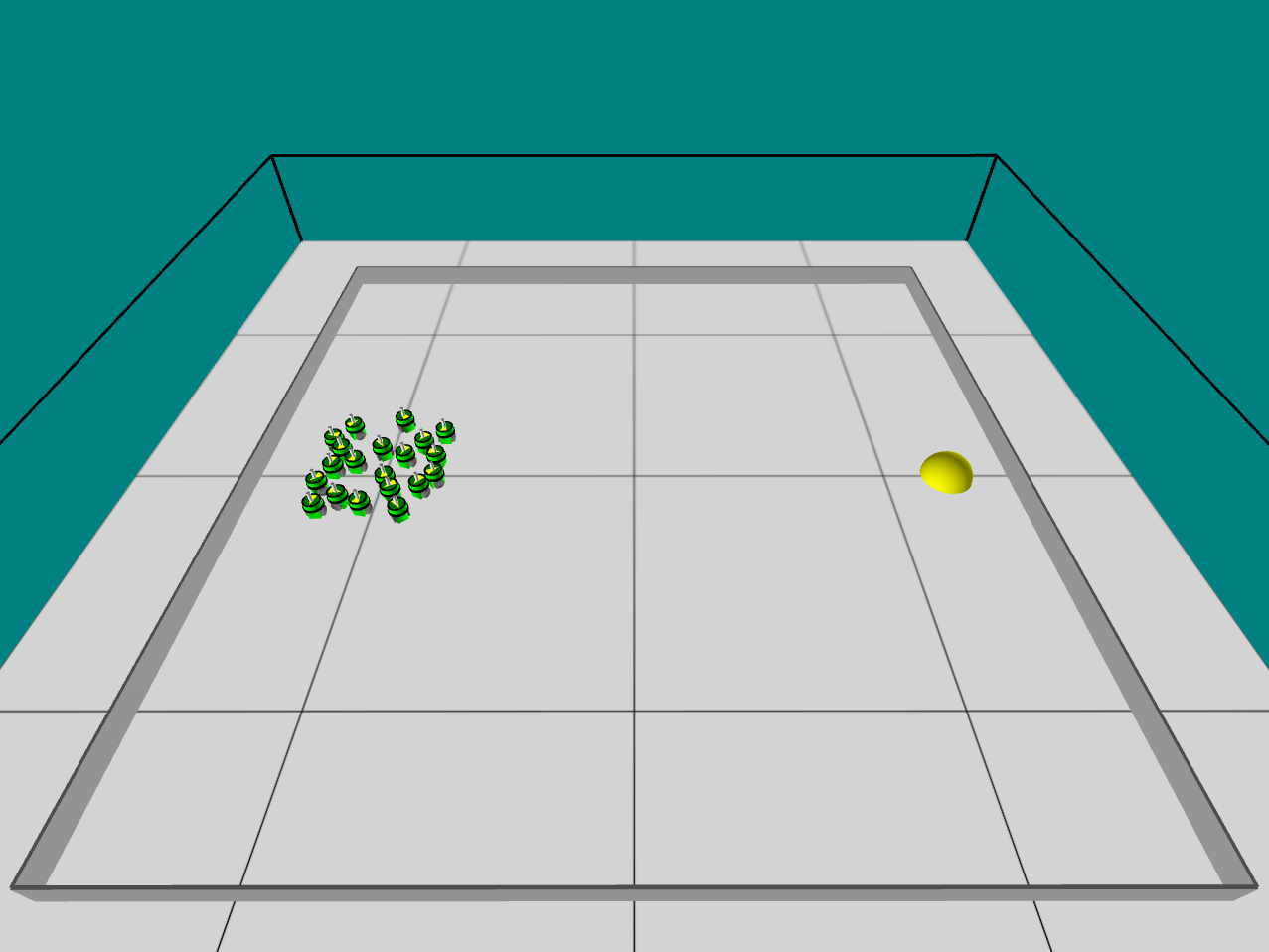}  %1.25
	}
	\subfloat[after $120$ $\unit{s}$]{
		\includegraphics[width = 1.1 in]{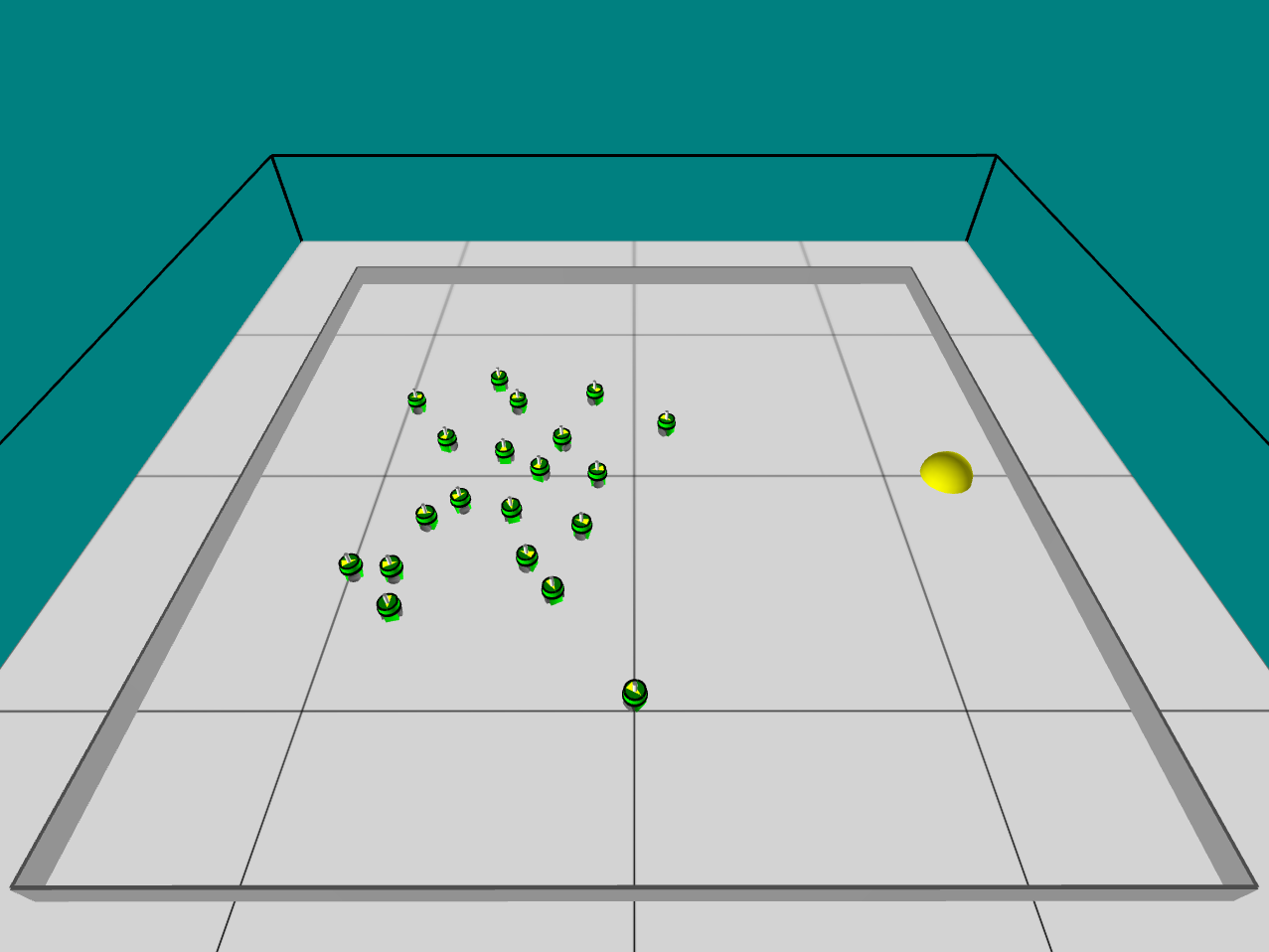}
	}
	\subfloat[after $240$ $\unit{s}$]{
		\includegraphics[width = 1.1 in]{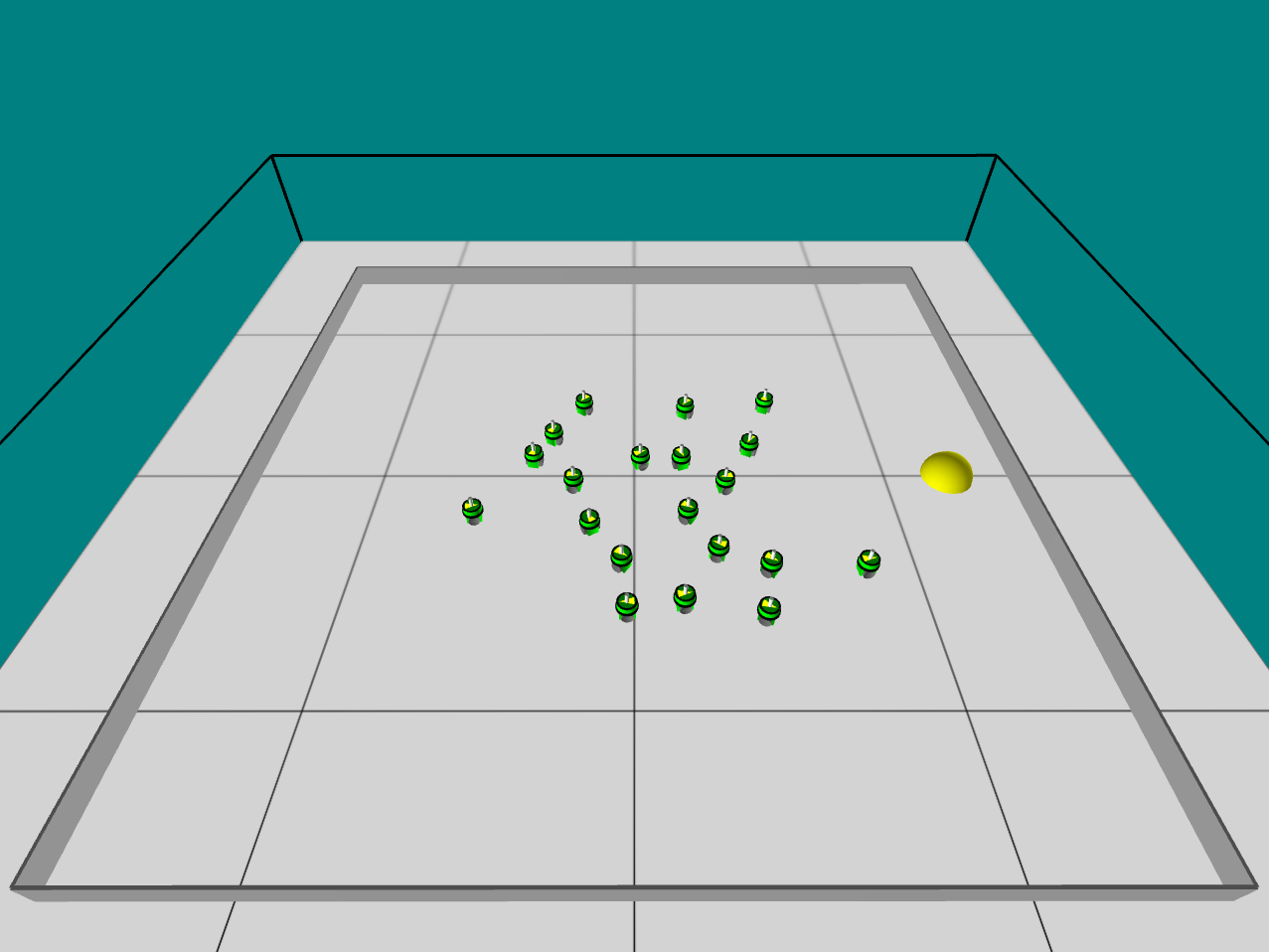}
	}
	\subfloat[after $400$ $\unit{s}$]{
		\includegraphics[width = 1.1 in]{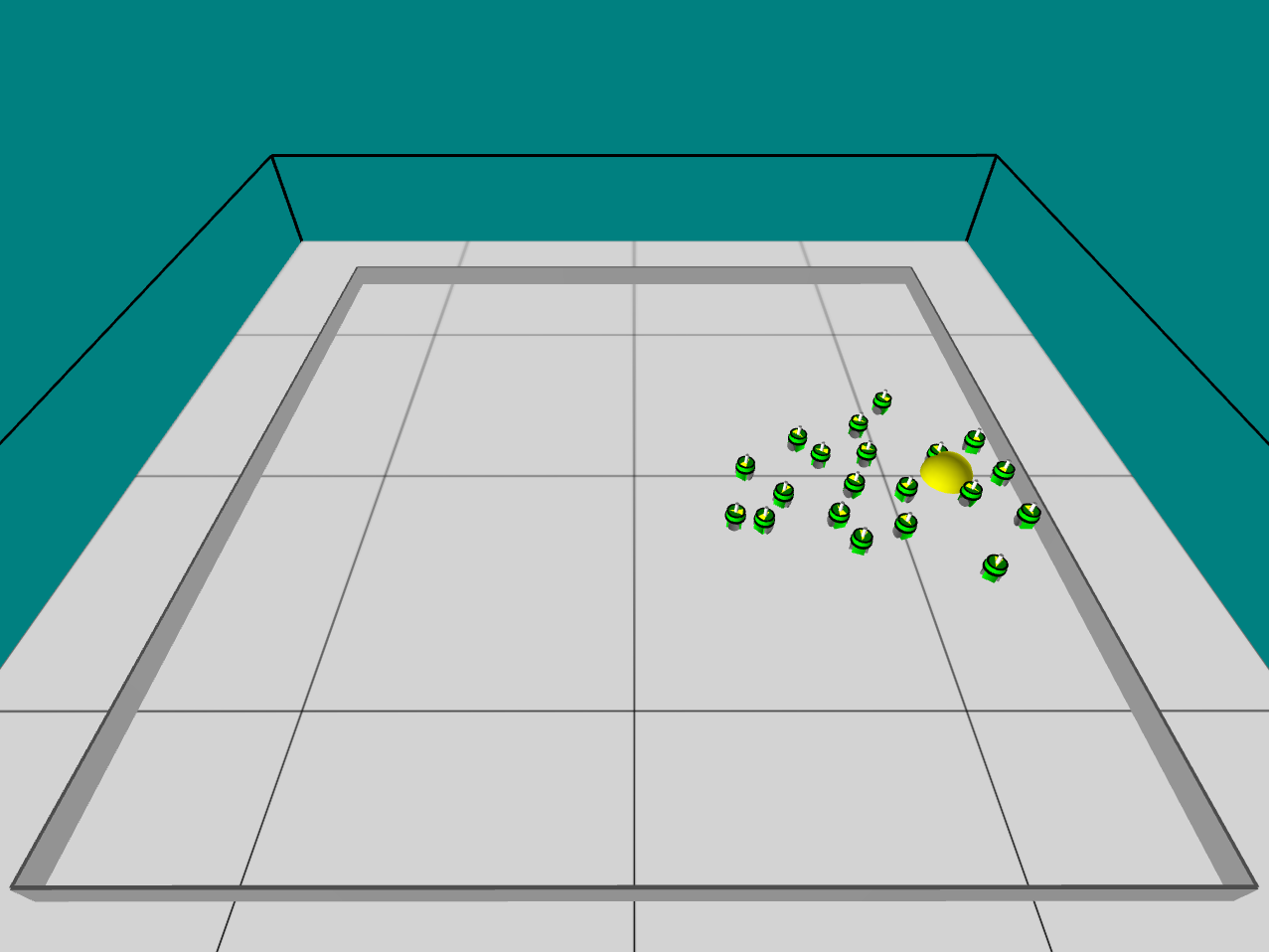}
	}\\
	\caption{Snapshots of the swarm taxis behaviour in simulation, using the automatically generated controller code from the RoboChart model. There are $20$ robots (green) and one beacon (yellow).}
	\label{fig:omega_algorithm_snapshoot}
\end{figure}

\subsubsection{Simulating the swarm taxis behaviour}

The swarm taxis behaviour is simulated in ARGoS~\citep{Pinciroli12}, which also has a built-in model of the e-puck. It is a 3D simulator. The simulated space can be divided into several sub-spaces that run different physics engines in parallel. The arena size is $\unit[400\times400]{cm^2}$. There is one beacon located in the right of the arena, and the robots are randomly initialized in the left region of the arena. Each robot is equipped with light sensors (to detect the beacon) and range-and-bearing sensors (to detect other robots nearby). The length of control step is set to $\unit[0.1]{s}$. Note that in the model shown in Figure~\ref{fig:omega_algorithm_controller_diagram} we did not attempt to optimize the value of each parameter (such as the time threshold and avoidance radius). 

We performed $10$ simulation trials with $20$ robots, and in each trial the robots can successfully move towards the beacon while maintaining a coherent group. Figure~\ref{fig:omega_algorithm_snapshoot} shows snapshots from a simulation trial, using the automatically generated controller code in RoboChart. A video showing the simulation of the two case studies and the automatically generated C++ code of the controllers can be found in the online supplementary materials~\citep{DARS2010Online}. 

%%This makes makes ARGoS suitable for simulated complex dynamic environments. 
%In both simulators, the speed of the left and right wheel of the e-puck can be set separately. 

\section{Conclusion}\label{sec:conclusion}

In this paper, we have presented a state-machine based framework RoboChart for modelling the controllers of autonomous robots, combined with the automatic generation of C++ code. We believe that this is the first framework that allows for both automatic code generation for robotic simulation, deployment and formal verification. The applicability of our approach has been demonstrated through modelling two case studies (self-organized aggregation and swarm taxis) in swarm robotics. The automatically generated code of the robot's controller was run in two different simulators, which again, demonstrates the flexibility of our approach.  

Our vision is to significantly reduce the gap between the high-level reasoning and low-level implementation through the use of formal methods and automatic code generation. The work presented can be seen as a first step towards the goal of verifying emergent behaviour, which is a potential application of our work to be investigated in the future. Our current focus is, however, to enrich the state machine specification of RoboChart by adding time and probability constructs, so that the framework can be applied to model a wide variety of robotic control systems. The formal semantics of RoboChart will also be enriched to make the verification feasible. In RoboChart, we focus on modelling the controller of a single robot, but we are investigating the possibility of using RoboChart models to simulate and analyse robotic swarms.
%In RoboChart, we focus on modelling the controller of a single robot, 
%Note that, this would not restrict our framework to be applied to multiple robotic systems.
%In the future, we will demonstrate the  power of RoboChart through comparing animation in CSP model and simulation with the automatically generated code to investigate whether the property verified in CSP is consistent with that shown in simulation.
%For example, to verify swarm behaviours, we will investigate the micro-level or macro-level controllers and try to predict the emergent behaviour of the swarm.    
%Note that, although the case studies presented in the paper are based on multiple robotic systems, the same modelling approach could also applied to single robotic systems. 

Currently, the generated controller code is a direct mapping from the elements in RoboChart to simulation. In the future, soundness of the simulation will be established by verifying the code generator. This can be realized using various software engineering techniques. In particular, we envisage that the CSP model generated from the RoboChart specification is a basis for establishing the correctness of the generated code using refinement. Practical verification can be carried out using a model checker like FDR (which also provides a facility to animate the model, and thus perform some validation), or using a theorem prover. 
%Besides, as we have the  CSP semantics of the model and an associated animation, we can compare the animation with the simulation to validate the code generator.

In this paper, we only automatically generate the code of controllers, however the simulation configurations (e.g. length of control step) in the case studies are manually defined. We intend to define simulations in an extended notation, from which the simulation configurations can also be specified. The simulation notation will be independent of specific programming languages such as C++ and Java, and of specific robotic platforms. 

Possible avenue for future work is the integration of RoboChart in other tools~\citep{Brambilla12, Francesca15}. For example, in~\citep{Francesca15}, an automatic design method was used to tune the free parameters of a predefined parametric architecture (e.g. probabilistic state machine) for the individual robot controller of a swarm. In this case, the controller architecture can be modelled in RoboChart, so that the obtained solution can be formally verified. In~\citep{Brambilla12}, a property-driven approach was proposed to design the controller of swarming robots. The designed controller can also be modelled in RoboChart to support both formal verification and code generation. 

Finally, we intend to model the environmental stimuli and generate code for physical robots. 
%In the later case, the semantics of RoboChart would enable verification not only in simulations but also on robotic hardwares.
%As the controller is implemented in a sequential way, we need to ensure that the length of the control step is small enough in order to capture the events.

%For future work, the code generator needs to know which simulator to be targeted, as the simulation configuration is platform dependent. 

%We wish to prove the emergent behaviour of the swarm system using the formal semantics. For example, in our case study of omega algorithm, an interesting property is to prove the minimum number of robots leading to the swarm taxis. 

%If the controller is implemented in a sequential way, we also need to assume that the length of the control step is small enough in order not to miss any events.

\section{Acknowledgement}\label{sec:acknowledgement}

The authors would like to acknowledge the support from EPSRC grant EP/M025756/1.

\bibliographystyle{abbrvnat}
\bibliography{referenc}

\end{document}